\documentclass[journal]{IEEEtran}

\usepackage{setspace}

\usepackage{amsmath, amsfonts}

\usepackage{cite}
\usepackage{algorithm}
\usepackage{algpseudocode}
\usepackage{enumerate}
\usepackage{color}
\usepackage{balance}
\usepackage{url}
\usepackage{multirow}  
\usepackage{pbox}  
\usepackage{flushend} 

\usepackage[pdftex]{graphicx}
\graphicspath{{./fig/}}%
\DeclareGraphicsExtensions{.jpg,.png,.pdf}
\usepackage[font={footnotesize}]{caption}
\usepackage[caption=false,font=footnotesize]{subfig}

\newcommand{\bi}{\begin{itemize}}	
\newcommand{\ei}{\end{itemize}}
\newcommand{\bn}{\begin{enumerate}}	
\newcommand{\en}{\end{enumerate}}
\newcommand{\bc}{\begin{center}}
\newcommand{\ec}{\end{center}}
\newcommand{\be}{\begin{equation}}
\newcommand{\ee}{\end{equation}}
\newcommand{\bea}{\begin{eqnarray}}
\newcommand{\eea}{\end{eqnarray}}
\newcommand{\ben}{\begin{equation*}}
\newcommand{\een}{\end{equation*}}
\newcommand{\beqa}{\begin{eqnarray}}
\newcommand{\eeqa}{\end{eqnarray}}
\newcommand{\btabu}{\begin{tabular}}
\newcommand{\etabu}{\end{tabular}}

\begin{document}
\title{Kernel Methods for Accurate UWB-Based Ranging with Reduced Complexity}

\author{Vladimir Savic, Erik G. Larsson, Javier Ferrer-Coll, and Peter Stenumgaard 
\thanks{Copyright (c) 2015 IEEE. Personal use of this material is permitted. However, permission to use this material for any other purposes must be obtained from the IEEE by sending a request to pubs-permissions@ieee.org. The original version of this manuscript is published in IEEE Transactions on Wireless Communications (DOI: 10.1109/TWC.2015.2496584).}%
\thanks{V. Savic and E. G. Larsson are with the Dept. of Electrical
  Engineering (ISY), Link\"{o}ping University, Sweden (e-mails:
  vladimir.savic@liu.se, erik.larsson@isy.liu.se). Javier Ferrer-Coll
  is with the Dept. of Electronics, Mathematics and Natural Sciences,
  University of G\"{a}vle, Sweden (e-mail:
  javier.ferrercoll@hig.se). Peter Stenumgaard is with the Swedish
  Defense Research Agency (FOI) (e-mail: peter.stenumgaard@foi.se).  }
\thanks{This work was supported by the project Cooperative
  Localization (CoopLoc) funded by the Swedish Foundation for
  Strategic Research (SSF), and Security Link. A part of this work was
  presented at IEEE SPAWC workshop, June 2014 \cite{Savic2014a}.
}%
}

\maketitle

\begin{abstract}
Accurate and robust positioning in multipath environments can enable
many applications, such as search-and-rescue and asset tracking. For
this problem, ultra-wideband (UWB) technology can provide the most
accurate range estimates, which are required for range-based
positioning. However, UWB still faces a problem with non-line-of-sight
(NLOS) measurements, in which the range estimates based on
time-of-arrival (TOA) will typically be positively biased. There are many techniques
that address this problem, mainly based on NLOS identification
and NLOS error mitigation algorithms. However, these techniques do not exploit
all available information in the UWB channel impulse
response. Kernel-based machine learning methods, such as Gaussian
Process Regression (GPR), are able to make use of all information, but
they may be too complex in their original form. In this paper, we
propose novel ranging methods based on kernel principal
component analysis (kPCA), in which the selected channel parameters
are projected onto a nonlinear orthogonal high-dimensional space, and a
subset of these projections is then used as an input for ranging. We
evaluate the proposed methods using real UWB measurements obtained in
a basement tunnel, and found that one of
the proposed methods is able to outperform state-of-the-art, even if little 
training samples are available.
\end{abstract}

\section{Introduction}\label{sec:intro}

UWB \cite{Gezici2005} is a promising technology for
range-based positioning in multipath environments. The large bandwidth
of a UWB signal provides a high temporal resolution for TOA-based
ranging, and allows propagation through thin obstacles. However, it
still faces a problem with NLOS measurements caused by thick obstacles
(i.e., when the direct path is completely blocked), in which case
TOA-based range estimates will typically be positively biased. This is especially a problem
in environments such as tunnels, warehouses, factories and urban
canyons. Consequently, a positioning system based on UWB range
estimates may provide poor performance unless careful action is taken. There are many techniques (see Section~\ref{sec:rel}) that address this problem. These techniques are mainly
based on NLOS identification and NLOS error mitigation algorithms, but
they do not exploit all available information in the UWB channel impulse
response. Kernel-based machine learning methods, such as GPR \cite{Wymeersch2012a}, are able
to make use of all information, but they may be too computationally
complex in their original form.

In this paper, we propose\footnote{This paper is a comprehensive extension of our conference paper \cite{Savic2014a}. New material includes: a detailed description of GPR for ranging, two new hybrid methods (kPCA+ and kPCA+GPR), and many additional experiments.} novel ranging methods
based on a nonlinear version of principal component analysis (PCA),
known as \textit{kernel} PCA (kPCA) \cite{Scholkopf1999}, in which the
selected channel parameters are projected onto a nonlinear orthogonal
high-dimensional space, and a subset of these projections is then used
as an input for ranging. Although popular for computer vision
problems, to the best of our knowledge, this technique has not so far
been used for UWB-based ranging. We evaluate the proposed methods
using real UWB measurements obtained in a basement tunnel of
Link\"{o}ping university. The most important conclusion is that one of the proposed methods, referred to as kPCA+GPR
(which makes combined use of kPCA and GPR techniques), is able to
outperform state-of-the-art. Moreover, kPCA+GPR is also much faster
than direct GPR-based ranging.

The remainder of this paper is organized as follows. In Section
\ref{sec:rel}, we provide an overview of state-of-the-art UWB-based
ranging techniques. In Section \ref{sec:toa-nlos}, we review the
generic approach for TOA-based ranging with NLOS identification and
error mitigation. Then, in Section \ref{sec:kernel}, we overview a GPR
technique for UWB-based ranging, and propose novel ranging
methods based on kPCA. Our experimental results, using UWB
measurements obtained in a tunnel environment, are provided in Section
\ref{sec:exp}. Finally, Section \ref{sec:conc} summarizes our
conclusions and provides some proposals for future work.

\section{Related Work}\label{sec:rel}

TOA estimates obtained from measured UWB impulse responses are typically biased in NLOS scenario. There are many proposals in the
literature for solving this problem \cite{Khodjaev2010}. These
techniques i) first attempt to distinguish between LOS and NLOS
conditions, and ii) then, if a NLOS condition is detected, they mitigate
the NLOS error. In what follows, we survey the available algorithms for
both these sub-problems.

\emph{1) NLOS identification.} This problem could be carried out by analyzing the variability
between consecutive range estimates \cite{Wylie1996,Borris1998},
relying on the assumption that NLOS measurements typically have much
larger variance than LOS measurements. However, this approach would
lead to very high latency since it requires a large number of
measurements. An alternative approach is to extract multiple
\emph{channel parameters} from the channel impulse response, and make a
NLOS identification based on these parameters. For example, in
\cite{Venkatesh2007a} three parameters are used (RMS delay-spread, TOA
and received signal strength (RSS)). This study
found that RMS delay spread is the most useful parameter for this
problem, but a combination of all three parameters can reduce the
misclassification rate. Another study \cite{Zhang2013} found that the
kurtosis provides valuable information for the NLOS identification. In
\cite{Marano2010}, a non-parametric least-square support-vector-machine
(LS-SVM) classifier is used. This approach does not require any
statistical model, but directly works with the training samples. A
non-parametric approach is also used in \cite{Gezici2003}, to construct
the probability density functions (PDFs) of the LOS and NLOS ranging
errors from training samples. Then, Kullback-Leibler divergence is
used to quantify the distance between these PDFs, and set the decision
threshold.

\emph{2) NLOS error mitigation.} Once NLOS identification is performed, one possibility is to simply discard all NLOS measurements, but this would lead to unnecessary loss of information. Therefore, NLOS error mitigation is
required to make NLOS measurements useful for ranging.  Since the
distribution of the NLOS error depends on the spatial distribution
of the scatterers, the mitigation could be performed by modeling
these scatterers \cite{Al-Jazzar2002,Al-Jazzar2002b}. However, this
approach is typically infeasible due to the complex geometry of
the environment, and the possible presence of dynamic obstacles. Another
way is to model the NLOS error as a function of some channel
parameter. For example, in \cite{Denis2003}, the authors found that
the NLOS error increases with the mean excess delay and the RMS
delay spread. Therefore, a simple empirical model can be used to
significantly reduce this error. 

In many cases, it may not be possible to detect an NLOS condition with
full certainty. In that case, a soft decision can be taken, that is, the
probability of NLOS condition is estimated. NLOS identification and
error mitigation are then combined into one single step \cite{Cong04},
and the ranging likelihood function becomes a mixture of LOS and NLOS
models. This kind of model is also obtained in \cite{Savic2014tw} by analyzing the measurements
from a tunnel environment. Finally, non-parametric kernel-based regression can be also
used to estimate the NLOS error as a function of multiple channel
parameters, without explicit NLOS identification. For this purpose,
GPR, LS-SVM regression, and relevance vector machines were used in 
\cite{Wymeersch2012a}, \cite{Marano2010}, and \cite{Nguyen2015}, respectively. These methods can achieve much
better ranging performance than standard TOA-based methods, but their
complexity is much higher.

\section{TOA-Based Ranging and NLOS Error Mitigation}\label{sec:toa-nlos}

Our goal is to estimate the range using the received UWB
signal. We first make the following assumptions:

\bi
\item We use a \textit{single} UWB measurement for ranging.
\item There are enough training UWB measurements (obtained offline or online, depending on the environment) for parameter estimation and learning.
\item The precise floorplan of the  area (especially, the positions of the scatterers and irregularities) is \textit{not available}, but there exist surfaces that cause \textit{multipath} propagation.
\item If the first path in the impulse response is \textit{undetectable}, the measurement is NLOS. Otherwise, even if the first path is not dominant \cite{Alsindi2007}, the measurement is LOS. This definition is reasonable because only an undetectable first path could lead to a serious ranging error \cite{Savic2014tw}.
\item The probability of NLOS condition is \textit{larger} than zero (otherwise, the ranging would be trivial).
\ei
Then, taking that we have measured the complex impulse responses
of the channel\footnote{Otherwise, if the output waveforms are available, a multipath extraction technique (such as CLEAN or ESPRIT) would be required.}, $h(t) = \sum\nolimits_{k} {{a_k}\delta (t - {\tau
    _k})}$ ($a_k \in \mathbb{C}$, $k=1,\ldots,N_k$, $N_k>1$, and ${\tau _k}$ is the delay of the
$k$-th path), we obtain its squared amplitude, $\left|h(t)\right|^2$, known as \textit{power delay profile} (PDP). However, since most of the components of the
PDP are typically caused by thermal noise, we consider only the
components above a certain threshold ${p_{TH}}$ [dBm], i.e.,
\be\label{eq:pdp-thresh} 
{p_h}(t) = \left\{ \begin{array}{l} {\left|
    {h(t)} \right|^2},\,{\rm{if}}\,\,{\left| {h(t)}
    \right|^2} >  {p_{TH}}\\ 0,\,\,\,\,\,\,\,\,\,\,\,\,\,{\rm{otherwise}}
\end{array} \right.
\ee
The threshold $p_{TH}$ is usually chosen empirically
\cite{Dardari2009,Guvenc2005,Savic2014tw} so as to satisfy desired
criteria (e.g, minimize the
false-alarm and the missed-detection rates). Then, we can extract a
number of \textit{channel parameters} from the PDP, such as TOA, RSS, RMS delay
spread \cite{Savic2014tw}. From these parameters, the range can be
determined. A naive way of computing the range estimate would be to take
$\hat{d}_{TOA}=c\tau_1$ where $\tau_1$ is estimated TOA, and $c \approx 3
\cdot 10^8$ m/s is the speed of light. However, this would typically
lead to a large positive error in the NLOS scenario. Thus, NLOS
identification and error mitigation are required to reduce this error.

In what follows we describe a general approach for estimating the
range from the channel parameters, which is essentially the basis
for the methods in \cite{Cong04,Savic2014tw}.  We define the binary
variable $H \in \{\rm{LOS}, \rm{NLOS}\}$, and consider the following
model: 
\be\label{eq:toa-bias} c{\tau _1} = \left\{ \begin{array}{l} d
  + {\mu _L } + {\nu
    _L},\,\,\,\,\,\,\,\,\,\,\,\,\,\,\,\,\,\,\,\,{\rm{if}}\,H =
  {\rm{LOS}}\\ d + g(\alpha_E) +
  {\nu_N},\,\,\,\,\,\,\,\,\,\,\,{\rm{if}}\,H = {\rm{NLOS}}
\end{array} \right.
\ee 
where $d$ is the true distance between transmitter and receiver,
${\nu_L}$ and ${\nu_N}$ are LOS and NLOS noise components (${\nu_L}
\sim p_L(\cdot)$, ${\nu_N}\sim p_N(\cdot)$), respectively, and ${\mu
  _L}$ is a known LOS bias (caused by finite bandwidth, false alarms,
or systematic errors). $g(\alpha_E)$ is the NLOS error modeled as a
(potentially nonlinear) function of some appropriately selected
channel parameter $\alpha_E$ (see Section \ref{sec:modelsel}). $\alpha_E$ should be empirically chosen 
so that it is strongly correlated with the NLOS error. In order to
make a soft decision on $H$, we use Bayes' rule:
\be\label{eq:nlos-bayes} p(H|{\alpha_I}) =
\frac{{p({\alpha_I}|H)p(H)}}{{\sum_ {H' \in \{\rm{LOS},
      \rm{NLOS}\}}p({\alpha_I}|H')p(H')}} \ee where $p(H)$ is the
prior, and $p(\alpha_I|H)$ is the likelihood function, and $\alpha_I$
is another channel parameter.  $\alpha_I$ should be empirically chosen
to be a good discriminator between LOS and NLOS. The prior on $H$ may
be chosen based on knowledge of the geometry of the area, or assumed
non-informative. The model for likelihood $p({\alpha_I}|H)$ in
(\ref{eq:nlos-bayes}) depends on the chosen channel parameter
$\alpha_I$ and can be obtained using training measurements.

The likelihood function for range estimation has a mixture form:
\beqa\label{eq:dist-lhood}
& p(\tau_1,\alpha_I,\alpha_E|d) \propto p(H = {\rm{LOS}}|{\alpha_I})p_L(c{\tau _1} - {\mu _L } - d) + \nonumber \\  
& ~~~~~~~~~~~~~~~~~p(H = {\rm{NLOS}}|{\alpha_I})p_N(c{\tau _1} - g({\alpha_E}) - d)
\eeqa
This (non-Gaussian) likelihood represents full statistical information
about unknown distance (assuming a non-informative prior), and can be used for Bayesian positioning
algorithms. Assuming that $p_L(\cdot)$ and $p_N(\cdot)$ are zero-mean
Gaussian (with variances $\sigma _L^2$ and $\sigma _N^2$), 
the minimum
mean-square-error (MMSE) estimate of the distance, and the
corresponding variance, are given by: 
\beqa\label{eq:dist-est-mitig} 
&\hat{d}_{TOA,M}=p(H = {\rm{LOS}}|{\alpha_I})\left(c{\tau _1} - {\mu
  _L}\right)+ \nonumber \\ 
& ~~~~~~~~~~~~~~~~p(H =
    {\rm{NLOS}}|{\alpha_I})\left(c{\tau _1} - g({\alpha_E})\right)
\eeqa 
\beqa\label{eq:var-mitig} 
&\sigma _{TOA,M}^2 =
    ~~~~~~~~~~~~~~~~~~~~~~~~~~~~~~~~~~~~~~~\nonumber \\ 
& p(H =
          {\rm{LOS}}|{\alpha _I}) \left( {{{(c{\tau _1} - {\mu _L} -
                {{\hat d}_{TOA,M}})}^2} + \sigma _L^2} \right) +
          \nonumber \\ 
& p(H = {\rm{NLOS}}|{\alpha _I})\left(
                    {{{(c{\tau _1} - g({\alpha _E}) - {{\hat
                              d}_{TOA,M}})}^2} + \sigma _N^2} \right)
\eeqa 
There are many variations of this approach available in literature. A
common (but sub-optimal) approach
\cite{Venkatesh2007a,Venkatesh2007,Guvenc2009} is to make a hard
decision on LOS/NLOS, and then estimate the distance. It is also
possible to use \textit{multiple} channel parameters for this problem \cite{Venkatesh2007a}, 
but this would likely lead to over-counting  of the same information since the
channel parameters are typically correlated \cite{Savic2014tw}. \footnote{Over-counting is a problem that occurs when
multiple correlated random variables are assumed to be uncorrelated. This problem typically leads to under-estimated variance.} The optimal solution for this problem would require a properly estimated joint distribution of the parameters which is a challenging problem in the presence of many parameters. Another way to solve this problem is by using kernel methods as will be shown in the following section.

\section{Kernel methods for ranging}\label{sec:kernel}

The goal is to perform ranging using all available channel parameters
from the PDP given by \eqref{eq:pdp-thresh}. Since some of the kernel methods
operate with centered and dimensionless data, we first transform the channel
parameters $\alpha_k$ ($k=1,\ldots,K$) as follows:
\be\label{eq:cent-norm}
a_k=\frac{\alpha_k-\mu_{\alpha_k}}{\sigma_{\alpha_k}} \ee 
where
$\mu_{\alpha_k}$ and $\sigma_{\alpha_k}$ are the mean and the standard
deviation of $\alpha_k$, respectively. Then, we gather all transformed
channel parameters into one vector ${\bf{a}}=(a_1,\ldots,a_K)^T$, that
will be used as an input for the kernel methods. In the
following subsections, we will first describe a state-of-the-art method based on GPR (given in \cite{Wymeersch2012a}), and then three novel methods based on kPCA.

\subsection{Gaussian process regression (GPR)}\label{sec:gp}

In this subsection, we describe  state-of-the-art in using GPR for range 
estimation, which will serve as a baseline in the
comparisons.  We consider the following regression problem:

\be\label{eq:nonlinReg} 
d=f({\bf{a}})+\omega 
\ee 
where $f(\cdot)$ is a
nonlinear function of the channel parameters, and $\omega$ is a
Gaussian random variable ($\omega \sim
\mathcal{N}(0,\sigma^2_\omega)$) that represents the error of
fitting. The problem can be solved by defining $f(\cdot)$ in a
parametric form, and learning its parameters. However, it is often
infeasible to decide which function is appropriate, so we assume that
this function is \textit{random}, and follows a \textit{Gaussian
  process (GP)} \cite{Rasmussen2006, Perez-Cruz2013}: $f(\cdot) \sim
\mathcal{GP}({\bf{m}},{\bf{K}})$ where ${\bf{m}}$ is a mean function
and ${\bf{K}}$ is a covariance function (also known as a
\textit{kernel} matrix). Note that it does not mean that the underlying process
is precisely Gaussian, but we can still use GP as a maximum entropy process,
for a given covariance function. For the problem at hand, we do not know the
mean function, so it is reasonable to assume\footnote{More precisely, 
if we do not know the mean function, we can assume a prior over it, and integrate it out. The resulting process will be zero-mean if this prior is symmetric around zero. For our problem, we do not have any information about the mean function, so we can assume a noninformative prior and set $p({\bf{m}})=\rm{const}$.} 
that ${\bf{m}}=0$. The kernel matrix ${\bf{K}}$ is used to model the correlation between
output samples as a function of the input samples.

Assuming that we have available a set of \textit{i.i.d.} training
samples ${{\cal T}_N} = \{ {d_n},{{\bf{a}}_n}\} _{n = 1}^N$ and a
single measurement ${\bf{a}}$, we would like to determine the
posterior density of the distance, i.e.,
$p(d|{\bf{a}},\mathcal{T}_N)$.  We also define the following sets:
${{\bf{f}}_N} = (f({{\bf{a}}_1}), \ldots ,\,f({{\bf{a}}_N}))$,
${{\bf{A}}_N} = ({{\bf{a}}_1}, \ldots ,\,{{\bf{a}}_N})$ and
${{\bf{d}}_N} = ({d_1}, \ldots ,\,{d_N})$. Taking into account the
previous assumptions and definitions, the likelihood function and the
GP prior over the training samples are, respectively, given by:
\be\label{eq:GPlhood} 
p({{\bf{d}}_N}|{{\bf{f}}_N}) = \prod\limits_{n =
  1}^N {p({d_n}|f({{\bf{a}}_n}))} = \prod\limits_{n = 1}^N
         {\mathcal{N}({d_n};f({{\bf{a}}_n}),\sigma _\omega ^2)} 
\ee
\be\label{eq:GPprior} 
p({{\bf{f}}_N}|{\bf{A}}_N) = {\cal N}({{\bf{f}}_N};0,{{\bf{K}}}) 
\ee
where $({{\bf{K}}})_{ij}=k({{\bf{a}}_i},{{\bf{a}}_j})$ ($ \forall {{\bf{a}}_i},{{\bf{a}}_j} \in \mathcal{T_N}$), $k(\cdot)$ is a kernel function, and ${\cal N}({\bf{x}};{\bf{\mu_x}},{\bf{\Sigma_x}})$ stands for a Gaussian distribution of the random variable ${\bf{x}}$ with  mean ${\bf{\mu_x}}$ and  covariance matrix ${\bf{\Sigma_x}}$. A widely used kernel function is a weighted sum of the squared exponential and linear terms, given by:
\be\label{eq:kernelsq}
k({{\bf{a}}_i},{{\bf{a}}_j}) = {\theta _0}{e^{ - {\theta _1}{{\left\| {{{\bf{a}}_i} - {{\bf{a}}_j}} \right\|}^2}}} + {\theta _2}{\bf{a}}_i^T{{\bf{a}}_j}
\ee
where the hyperparameters $\boldsymbol\theta=(\theta _0,\theta _1,\theta _2)$, along with $\sigma_{\omega}$, can be estimated from the training data by maximizing the log-marginal likelihood $\log{p({{\bf{d}}_N}|{{\bf{A}}_N},{\bf{\boldsymbol\theta}},\sigma_{\omega})}$. This can be achieved numerically using some gradient-based optimization technique (more details in \cite{Rasmussen2006}). With this kernel, the correlation between the output samples is higher if the Euclidean distance between the corresponding input samples is smaller. Note that the standard linear regression ($f({{\bf{a}}})={\bf{w}}^T{{\bf{a}}}$) is obtained as a special case by setting $\boldsymbol\theta=(0,0,1)$.

Now, we extend the training sample set with a test measurement ${{\bf{a}}}$, and define the \textit{extended} GP prior:
\be\label{eq:GPextPrior}
p(f({\bf{a}}),{{\bf{f}}_N}|{\bf{a}},{{\bf{A}}_N}) = {\cal N}\left(\left[ \begin{array}{l}
~{{\bf{f}}_N}\\
f({\bf{a}})
\end{array} \right]; {0,\left[ {\begin{array}{*{20}{c}}
{\bf{K}}&{\bf{k}}\\
{{{\bf{k}}^T}}&{k({\bf{a}},{\bf{a}})}
\end{array}} \right]} \right)
\ee
where ${\bf{k}} = {\left[ {k({{\bf{a}}_1},{\bf{a}}),\,k({{\bf{a}}_2},{\bf{a}}),\, \ldots ,k({{\bf{a}}_N},{\bf{a}})} \right]^T}$. To include the information from the likelihood, we apply Bayes' rule:
\be\label{eq:bayesGP}
p(f({\bf{a}}),{{\bf{f}}_N}|{\bf{a}},{{\cal T}_N}) \propto p({{\bf{d}}_N}|{{\bf{f}}_N})p(f({\bf{a}}),{{\bf{f}}_N}|{\bf{a}},{{\bf{A}}_N})
\ee
where we used the fact (see eq. \eqref{eq:nonlinReg}) that ${{\bf{d}}_N}$ is independent of $\{{\bf{a}},{{\bf{A}}_N}\}$, given ${{\bf{f}}_N}$. Finally, we find the desired posterior by marginalizing over the latent function:
\be\label{eq:postMarg}
p(d|{\bf{a}},{{\cal T}_N}) = \int {\int {p(d|f({\bf{a}}))} } p(f({\bf{a}}),{{\bf{f}}_N}|{\bf{a}},{{\cal T}_N})d{{\bf{f}}_N}df({\bf{a}})
\ee
where $p(d|f({\bf{a}})) = {\cal N}(d;f({\bf{a}}),\sigma _\omega ^2)$, and $\int {\int{}} $ denotes $N+1$ integrations. Since all distribution inside the integral are Gaussian, the posterior is also Gaussian, so $p(d|{\bf{a}},{{\cal T}_N}) = {\cal N}\left(d; {{\mu _{GPR}},\sigma _{GPR}^2} \right)$ with the parameters given by:
\be\label{eq:postMean}
{\mu _{GPR}} = {{\bf{k}}^T}{({\bf{K}} + \sigma _\omega ^2{{\bf{I}}_N})^{ - 1}}{{\bf{d}}_N}
\ee
\be\label{eq:postVar}
\sigma _{GPR}^2 = \sigma _\omega ^2 + k({\bf{a}},{\bf{a}}) - {{\bf{k}}^T}{({\bf{K}} + \sigma _\omega ^2{{\bf{I}}_N})^{ - 1}}{\bf{k}}
\ee
The MMSE estimate of the distance is given by $\hat{d}_{GPR}={\mu _{GPR}}$ and the remaining uncertainty by the variance $\sigma _{GPR}^2$. Therefore, GPR is capable of providing full statistical information (i.e., both the mean and the variance). This stands in contrast to other machine learning methods such as SVM.\footnote{A variation of SVM can provide a measure of uncertainty, but in an ad-hoc way, as pointed out in \cite[Section 6.4]{Rasmussen2006}. GPR provides the best possible estimate in the Bayesian sense (by finding the posterior using Bayes rule). } Another important characteristic of GPR is that $\sigma _{GPR}$ is small in the areas where the training samples lie, and large in the areas with no (or few) training samples. Finally, assuming offline training, we can precompute the inverse of the matrix ${{\bf{K}} + \sigma _\omega ^2{{\bf{I}}_N}}$ and store into  memory,\footnote{This computation requires $\mathcal{O}(N^3)$ operations, but it is done only once in this case.} so the complexity of GPR is $\mathcal{O}(N^2)$.

Note that there is a minor difference between the original method in \cite{Wymeersch2012a} and the algorithm above. In \cite{Wymeersch2012a}, GPR is used to estimate the bias of the TOA-based range estimate, whereas the above
method directly estimates the range.

\subsection{Kernel Principal Component Analysis (kPCA)}\label{sec:kpca}

The GPR method, described in Section~\ref{sec:gp}, provides the optimal estimate
\cite{Perez-Cruz2013} of the distance in the MMSE sense under
  the assumption that the unknown function $f(\cdot)$ follows a
  GP. However, in some applications it may be too computationally
complex, especially, if there are many training samples. In this
section, we describe an alternative approach based on kPCA,
which is proposed in a preliminary form in \cite{Savic2014a}.

\begin{figure}[!t]
\centerline{
\includegraphics[width=1.07\columnwidth]{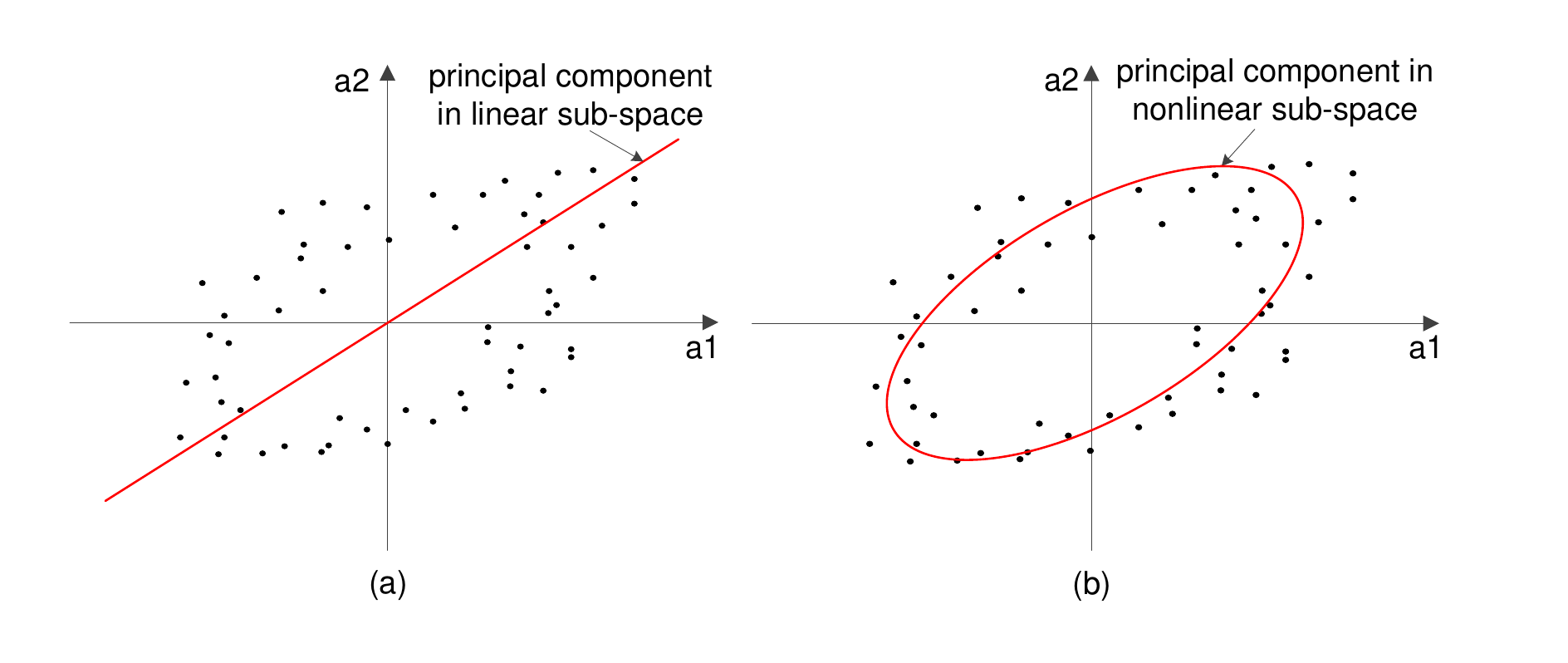}}
\caption{Principal components of 2D data in (a) linear and, (b) nonlinear sub-space. The latter one is much more accurate projection.}
\label{fig:pca-example}
\end{figure}

Instead of directly using the raw channel parameters ${\bf{a}}$, one
possible approach could be to apply eigenvalue decomposition to
decorrelate them, and then retain $M$ largest eigenvalues and
corresponding eigenvectors, that can be used for ranging. This
procedure is known as \textit{principal component analysis}
(PCA). However, PCA projects the data in a linear sub-space, so it is
not useful if the data lies on a nonlinear manifold. An example for 2D
data is shown in Fig. \ref{fig:pca-example}. Therefore, we propose to
apply \textit{kernel} PCA (kPCA) initially developed in its general
form in \cite{Scholkopf1999}. kPCA differs from PCA in that it 
projects the data to an arbitrary \emph{nonlinear} manifold. kPCA is already successfully
applied to pattern recognition from images \cite{Scholkopf1999}, and it is found that nonlinear principal components can provide
much better recognition rates than corresponding linear principal components.

Consider a nonlinear transformation $\boldsymbol\phi({\bf{a}})$, which
transforms $K$-dimensional vector ${\bf{a}}$ to an $N$-dimensional vector
in a feature space. The feature space has large (possibly, infinite)
dimension ($N>>K$). For now, we also assume that
$\boldsymbol\phi({\bf{a}})$ has zero mean, so that its $N \times N$
covariance matrix is given by
${\bf{C}}=E(\boldsymbol\phi({\bf{a}})\boldsymbol\phi({\bf{a}})^T)$. Then,
we could find the principal components using eigenvalue expansion of
${\bf{C}}$, but this is not possible to compute explicitly since the
feature space has high dimension. However, it can be shown
\cite[Chapter 12]{Bishop2006} that eigenvalue decomposition depends on
$\boldsymbol\phi({\bf{a}})$ only via the inner product
$k({\bf{a}},{\bf{a}}_n)=\boldsymbol\phi({\bf{a}})^T\boldsymbol\phi({\bf{a}}_n)$
(where ${\bf{a}}_n$ is a training sample, i.e., ${\bf{a}}_n \in
\mathcal{T}$), known as a \textit{kernel} function. Thus, we find
principal components using the following eigenvalue expansion:
 \be\label{eq:eig-kernel}
{\bf{K}}{{\bf{v}}_n} = {\lambda _n}N{{\bf{v}}_n}
\ee
where ${\bf{K}}$ is an $N \times N$ kernel matrix\footnote{Although we use the same symbol for the kernel matrix in GPR and kPCA, they represent the different quantities.} with $({{\bf{K}}})_{ij}=k({{\bf{a}}_i},{{\bf{a}}_j})$, $ \forall {{\bf{a}}_i},{{\bf{a}}_j} \in \mathcal{T}.$
The eigenvalues and eigenvectors of ${\bf{K}}$ are given by $\lambda _nN$ and ${{\bf{v}}_n}$ ($n=1,\ldots,N$), respectively. Then, the projection of a test point ${\bf{a}}$ onto eigenvector $i$ is given by
\be\label{eq:projection}
{y_i}({\bf{a}}) = \sum\limits_{n = 1}^N {{v_{in}}k({\bf{a}},{{\bf{a}}_n})}
\ee
Therefore, we can obtain all projections using kernel functions, without explicit work in feature space. In practice, the feature vector is not zero-mean, but it only means that we need to replace ${\bf{K}}$ with a \textit{Gram} matrix ${\bf{\tilde K}}$:
\beqa
\label{eq:gram}
&{\bf{\tilde K}} = {\bf{K}} - {{\bf{1}}_N}{\bf{K}} - {\bf{K}}{{\bf{1}}_N} + {{\bf{1}}_N}{\bf{K}}{{\bf{1}}_N}
\eeqa
where ${{\bf{1}}_N}$ is the $N \times N$ matrix with all values equal to $1/N$. We see that this matrix can be also computed without the feature vectors. Therefore, for kPCA, we only need to define a kernel function. Popular choices \cite{Scholkopf1999} are Gaussian, sigmoid and polynomial kernels. Since we know that the range is a linear function of the TOA in case of LOS, and possibly a nonlinear function of all features in case of NLOS, a kernel that can model both linear and nonlinear relationships is appropriate. Therefore, we choose a polynomial kernel, given by:
\be
\label{eq:poly-kernel}
k({\bf{a}},{{\bf{a}}_n}) = ({{\bf{a}}^T{{\bf{a}}_n}}+1)^c
\ee
where $c \in \mathbb{N}$ is the degree of the polynomial. The degree $c$ can be found empirically. Note that the standard PCA is a special case that results if we select the linear kernel $k({\bf{a}},{{\bf{a}}_n})={\bf{a}}^T{{\bf{a}}_n}$.

Then, given $M$ principal components ${y_i}({\bf{a}})$ corresponding to the $M$ largest eigenvalues, we need to provide a model that will relate them with the unknown range. Since this relationship is unknown in general, we use a simple linear model. A nonlinear model would be  redundant anyway since we already had the opportunity to perform arbitrary nonlinear transformations using kPCA. Therefore, we assume that the projected features ${y_i}={y_i}({\bf{a}})$ are linear functions of the true distance:
\be
\label{eq:proj-distance}
{y_i} = {b_{1,i}}d + {b_{0,i}} + {\nu _{{y,i}}}
\ee
where ${b_{1,i}}$ and ${b_{0,i}}$ are parameters that can be found using least-squares curve fitting (using the same training samples, $\mathcal{T}$), and ${\nu _{{y,i}}}$ is a noise component. Since the distribution of the noise is also unknown, we assume that it is zero-mean Gaussian as the distribution that maximizes the entropy (given the variance $\sigma_{y,i}^2$, which can be found from the training samples). Assuming that we retain $M$ principal components, and taking into account that they are orthogonal to each other, the likelihood function is given by:
\beqa
\label{eq:lhood-kpca}
& p({y_1}, \ldots ,{y_M}|d) = \prod\limits_{i = 1}^M {p({y_i}|d)}  = \nonumber \\
& ~~~~~~~~\prod\limits_{i = 1}^M {\mathcal{N}({y_i};{b_{1,i}}d + {b_{0,i}},\sigma _{y,i}^2)}  = \nonumber \\
& \prod\limits_{i = 1}^M {\mathcal{N}\left( {d;\frac{{{y_i} - {b_{0,i}}}}{{{b_{1,i}}}},\frac{{\sigma _{y,i}^2}}{{b_{1,i}^2}}} \right)} 
\eeqa
The maximum  number of principal components is equal to the total number of the training samples ($N$), which is much higher than the dimensionality of the original data ($K$). Since $M$ is expected to be larger than $K$, the main purpose of kPCA is nonlinear feature extraction, in contrast to PCA, which is typically used for dimensionality reduction. 

From \eqref{eq:lhood-kpca}, we can find the  MMSE estimate of the distance, 
\be\label{eq:dist-est-kpca}
{\hat d_{kPCA}} = \sigma^2_{kPCA}\sum\limits_{i = 1}^M {\left( {{b_{1,i}}\frac{{{y_i} - {b_{0,i}}}}{{\sigma _{y,i}^2}}} \right)}
\ee
where
\be\label{eq:var-est-kpca}
\sigma^2_{kPCA}={\left( {\sum\limits_{i = 1}^M {\frac{{b_{1,i}^2}}{{\sigma _{y,i}^2}}} } \right)^{ - 1}}
\ee
is the variance of the estimate.
Recall that this estimate is valid for both LOS and NLOS scenarios, so NLOS identification is not needed. Assuming that eigenvalue decomposition is done offline, the complexity of kPCA is $\mathcal{O}(MN)$ since we need to find only $M$ principal components using \eqref{eq:projection}. Therefore, kPCA is approximately $N/M$ times faster than GPR. However, this approach will provide a sub-optimal result since the noise in \eqref{eq:proj-distance} is not Gaussian in the general case.

\subsection{Hybrid methods: kPCA+ and kPCA+GPR}\label{sec:kpcagp}
One problem of the  kPCA and GPR methods described above is that they do not exploit the fact that some
measurements will be LOS, and in that case  the range can be directly obtained from the TOA. Hence,  the kernel methods are only needed to handle NLOS measurements. However, since it may not be possible to make a reliable LOS/NLOS identification (i.e., estimate $H$), we need to use a soft-decision approach in a similar way as in Section \ref{sec:toa-nlos}.

Since kPCA provides us many (up to $N$) projected features $y_i$, it is logical to assume that some of them will provide us useful information about  $H$. However, the projected features represent a complex and unknown function of the raw channel parameters, so it is not easy to determine the precise model for the distribution $p(y_i|H)$. Therefore, using the same principle as in Section~\ref{sec:kpca}, we approximate this distribution with a Gaussian as follows:
\be\label{eq:lhood-proj}
p({y_i}|H) = \left\{ \begin{array}{l}
{\cal N}({y_i};{\mu _{L,{y_i}}},\sigma _{L,{y_i}}^2),\,\,\,\,\,{\rm{if}}\,\,\,H = {\rm{LOS}}\,\,\\
{\cal N}({y_i};{\mu _{N,{y_i}}},\sigma _{N,{y_i}}^2),\,\,\,\,{\rm{if}}\,\,\,H = {\rm{NLOS}}
\end{array} \right.
\ee
where the parameters ${\mu _{L,{y_i}}}$, $\sigma _{L,{y_i}}$, ${\mu _{N,{y_i}}}$, $\sigma _{N,{y_i}}$ are found from the training set. This approximation will cause some loss of information, but we are now able to use multiple uncorrelated features, in contrast to the approach described in Section \ref{sec:toa-nlos}. Consequently, the total likelihood function is given by:
\be\label{eq:lhood-proj-all}
p({y_1}, \ldots ,{y_{M'}}|H) = \prod\limits_{i = 1}^{M'} {p({y_i}|H)}
\ee
where $M'$ is the number of retained features for this classification problem (not necessarily equal to $M$). Then, the posterior distribution of $H$ is given by
\be\label{eq:post-proj-all}
p(H|{y_1}, \ldots ,{y_{M'}}) \propto p({y_1}, \ldots ,{y_{M'}}|H)p(H)
\ee
where $p(H)$ is the prior distribution of $H$. Finally, the MMSE estimate of the distance is given by:
\beqa\label{eq:dist-est-kpca+}
& \hat{d}_{kPCA+}=p(H = {\rm{LOS}}|{y_1}, \ldots ,{y_{M'}})\left(c{\tau _1} - {\mu _L}\right)+ \nonumber \\
& ~~~~~~~p(H = {\rm{NLOS}}|{y_1}, \ldots ,{y_{M'}})\hat{d}_{kPCA}
\eeqa
where $\hat{d}_{kPCA}$ is the kPCA estimate of distance found by \eqref{eq:dist-est-kpca}, but with one difference: the training set consists \textit{only} of NLOS samples (the number of these samples is denoted by $N_N$). The variance of this estimate can be computed in the same fashion as the one in \eqref{eq:var-mitig}. Assuming that both eigenvalue decompositions are done offline, the complexity of this approach, referred to as kPCA+, is $\mathcal{O}(M'N+MN_N)$.

Alternatively, we may use GPR to estimate the distance from the NLOS training samples. In that case, the distance estimate is given by:
\beqa\label{eq:dist-est-kpca+}
& \hat{d}_{kPCA+GPR}=p(H = {\rm{LOS}}|{y_1}, \ldots ,{y_{M'}})\left(c{\tau _1} - {\mu _L}\right)+ \nonumber \\
& ~~~~~~~p(H = {\rm{NLOS}}|{y_1}, \ldots ,{y_{M'}})\hat{d}_{GPR}
\eeqa

The complexity of this approach, referred to as kPCA+GPR, is $\mathcal{O}(M'N+N_N^2)$. With a reasonable assumption that $N_N<<N$, kPCA+GPR would have a similar complexity as kPCA+, and would be much faster than original GPR.

A possible problem of the proposed kernel algorithms is the training phase, in which we need to collect enough measurements and corresponding ranges. This training can be done either offline or online, depending how variable the channel is in the considered environment. Since our goal is ranging using UWB signals (which can pass through thin obstacles \cite{Savic2014tw}), the correct TOA can be estimated as long as the first path is detectable (even attenuated). Therefore, if the variability in the environment is caused by people walking or other thin objects (the required level of thinness depends on the signal bandwidth), the training can be done offline, as in the environment considered in the next section. Otherwise, e.g., in the presence of moving vehicles and other machinery, the training should be done online. In that case, a set of pre-deployed anchors on known (manually recorded) positions should be used. Their positions should be carefully chosen (e.g., uniformly deployed) in order to provide a sufficient statistics for training of parameters. This training can be repeated periodically (with a lower frequency than the online algorithm), or manually triggered once some change in the environment is detected. Since this procedure may be too slow in larger areas, the online training should be done only for the parts of the area in which the environmental changes are expected.

\section{Experimental Results}\label{sec:exp}

In this section, we analyze the performance of the proposed ranging methods using UWB measurements obtained in a basement tunnel of Link\"{o}ping university (LiU) in Sweden.

\subsection{Measurement Setup and Scenarios}\label{sec:setup}

\begin{table}[!tb]
\caption{Measurement parameters}\label{table:param}
\centering
\begin{tabular}{l || l}
Parameter & Value \\
\hline\hline
Signal power & 12 dBm \\
Waveform & sinusoidal sweep\\
Center frequency & 3.5 GHz\\
Bandwidth & 2 GHz\\
Number of freq. points & 3001\\
Sweep time & 263 ms\\
Resolution & 0.5 ns (15 cm)\\
Antenna range & 1.71 - 6.4 GHz\\
Antenna gain & 5 - 7.5 dBi\\
Cable attenuation & 0.65 dB/m\\
\hline \hline
\end{tabular}
\end{table}

\begin{figure}[!t]
\centerline{
\includegraphics[width=0.85\columnwidth]{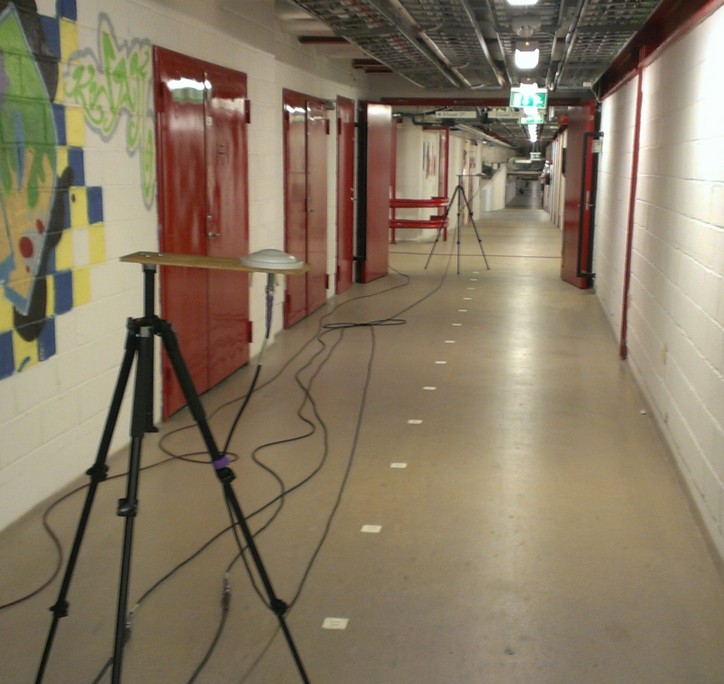}}
\caption{Illustration of LOS experiments in LiU tunnel.}
\label{fig:los-exper}
\end{figure}

\begin{figure*}[!tb]
\centerline{
\includegraphics[width=1\textwidth]{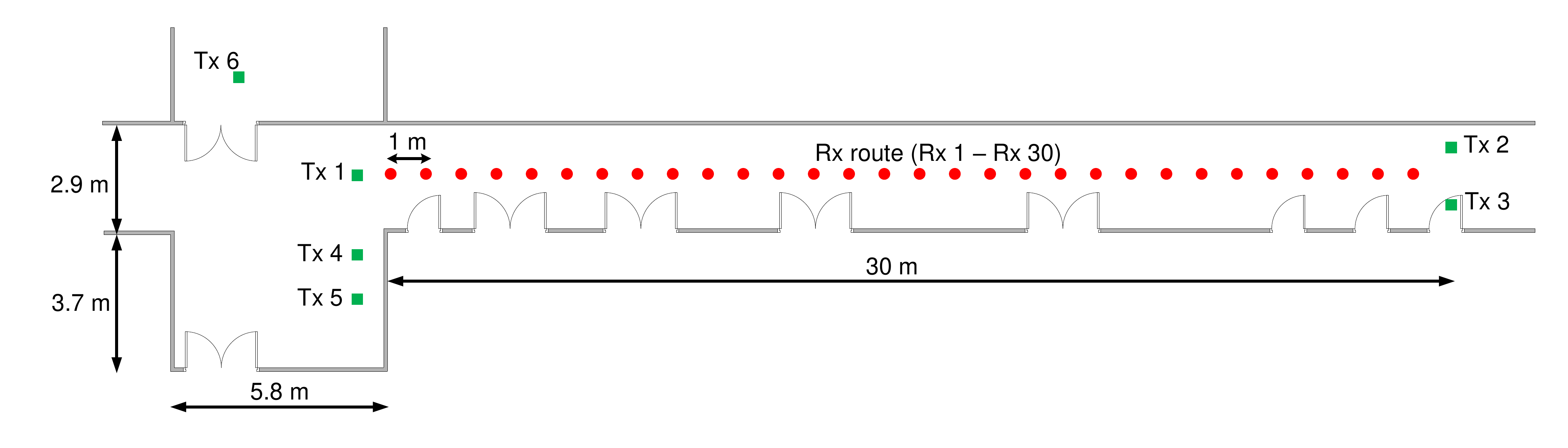}}
\caption{Deployment of transmitters (Tx) and receivers (Rx) in the LiU tunnel. There are 6 transmitter positions (marked with green squares), and 30 receiver positions (marked with red circles). Transmitters Tx 1-3 are used to test the LOS and two NLOS scenarios (by putting a metal sheet and a person in front of them), while transmitters Tx 4-6 are used to test NLOS caused by tunnel wall. The height of the tunnel is 2.8 m.}
\label{fig:floorplan-liu}
\end{figure*}

For our experiments, the measurement setup consisted of a vector
network analyzer (VNA), two UWB omni-directional antennas and coaxial
cables to connect the antennas with the VNA. A personal computer is
used to set the VNA parameters and extract the multiple frequency
responses from the VNA. We used a swept-frequency sinusoidal signal
(with 3001 points) to characterize the channel between 2.5 and 4.5
GHz. The power level was set to 12 dBm, and a calibration of the
system is performed to compensate for the effects of VNA, cables and
antennas. Then, the frequency responses are transferred to the PC
where a Hann window is used to reduce the out-of-band noise, and to
ensure causality of the time-domain responses. Finally, by applying
the Inverse Fast Fourier Transform (IFFT), the complex impulse
responses are estimated, and then, PDPs are calculated. We summarize
the main parameters in Table \ref{table:param}.

Using the described setup, the measurements have been carried out in a basement tunnel of LiU (Fig. \ref{fig:los-exper}). The tunnel walls, excluding the metal doors, are built of concrete blocks with steel reinforcement. The ceiling is also made of concrete, but with many metal pipes. Four different scenarios have been considered: LOS, and NLOS caused by three different obstacles: metal sheet, person, and tunnel wall. For each of these scenarios, we placed the transmitter (Tx) in 3 positions and receiver (Rx) in 30 positions forming the route through the tunnel, as shown in Fig. \ref{fig:floorplan-liu}. For each Tx-Rx pair, we obtained 10 PDPs, so we obtained 3600 PDPs in total (900 per scenario). Since our initial analysis showed that thin obstacles (metal sheet and person) provide results similar as LOS (i.e., the direct path is detectable), we will consider them as LOS in the analyses. Therefore, we have available 2700 LOS samples and 900 NLOS samples. Half of these samples (1350 LOS samples and 450 NLOS samples) will be used as training data, and the rest will be used to test the ranging performance. Note that the training samples should include all the scenarios that are expected to be encountered in the online phase (i.e., LOS and all types of NLOS).\footnote{We also tested other appropriate training/test splits, and found that they provide nearly the same results.}

\subsection{Model Selection}\label{sec:modelsel}

\begin{table}[!tb]
\caption{The parameters estimated from the training samples.}
\label{table:ranging-param-toa}
\centering
\begin{tabular}{c||c}
 Parameters &  Estimated values \\
 \hline\hline
$\sigma_{L}$ & 0.16 m \\
$\sigma_{N}$ & 1.61 m \\
$\lambda_{L}$ & 0.333 ns$^{-1}$ \\
$\lambda_{N}$ & 0.075 ns$^{-1}$\\
$(p_2, p_1, p_0)$ & $(0.00087,-0.2, 11.72)$ \\
$\boldsymbol\theta$ & $(64.6, 0.57, 1.59)$ \\
$\sigma_{\omega}$ & 0.5 \\\hline\hline
\end{tabular}
\end{table}

We first extracted many parameters from the available PDPs (i.e., all parameters that have some physical meaning). Then, we tested the cross-correlation between each pair of parameters, and removed those that do not provide any additional information. From the remaining subset, we removed those which have negligible correlation with both the true range and the NLOS bias. The remaining $K=8$ channel parameters are used for kernel methods: \textit{TOA, RSS, maximum received power, mean excess delay, maximum excess delay, RMS delay spread, rise time and kurtosis} (their definitions are available in \cite[Section IV]{Savic2014tw}). This parameter set is very similar to the sets \cite{Denis2003, Marano2010, Wymeersch2012a,Nguyen2015} widely used for ranging in indoor environments.
For TOA with NLOS identification and error mitigation, we found that rise time is the
best channel parameter ($\alpha_I$) for NLOS identification, and
maximum excess delay is the best channel parameter ($\alpha_E$) for
NLOS error mitigation. Regarding the model in \eqref{eq:toa-bias}, our
measurements indicated that $\nu_L$ and $\nu_N$ approximately follow a
zero-mean Gaussian distribution (with variances $\sigma _L^2$ and $\sigma _N^2$), while $g(\alpha_E)$ can be modeled
with a second-order polynomial function (i.e.,
$g({\alpha_E})=p_2{\alpha_E}^2+p_1{\alpha_E}+p_0$). The likelihood
function $p({\alpha_I}|H)$ is assumed to follow an exponential
distribution with different decay rates in the LOS and NLOS scenarios (given
by $\lambda_{L}$ and $\lambda_{N}$, respectively), and the prior
$p(H)$ is assumed to be non-informative. The justification of all
these models is available in \cite[Section V]{Savic2014tw}. For the
baseline GPR method, we used the Matlab toolbox (available online \cite{GPMLweb}) to
obtain $\boldsymbol\theta$ and $\sigma_{\omega}$, and to perform the
regression. The numerical values of all parameters are shown in Table
\ref{table:ranging-param-toa}. We note the following: i) the standard deviation of the range estimate is much higher for the NLOS scenario ($\sigma _N >> \sigma _L$), ii) the LOS rise time decays much faster than the NLOS rise time ($\lambda_{L} >> \lambda_{N}$), and iii) the squared exponential term in the kernel, given by \eqref{eq:kernelsq}, has much higher influence than the linear term ($\theta_0>>\theta_2$). Note that these parameters are valid only for this specific environment, which means that the training phase should be repeated for the other environments.

\begin{figure}[!tb]
\centerline{
\includegraphics[width=1\columnwidth]{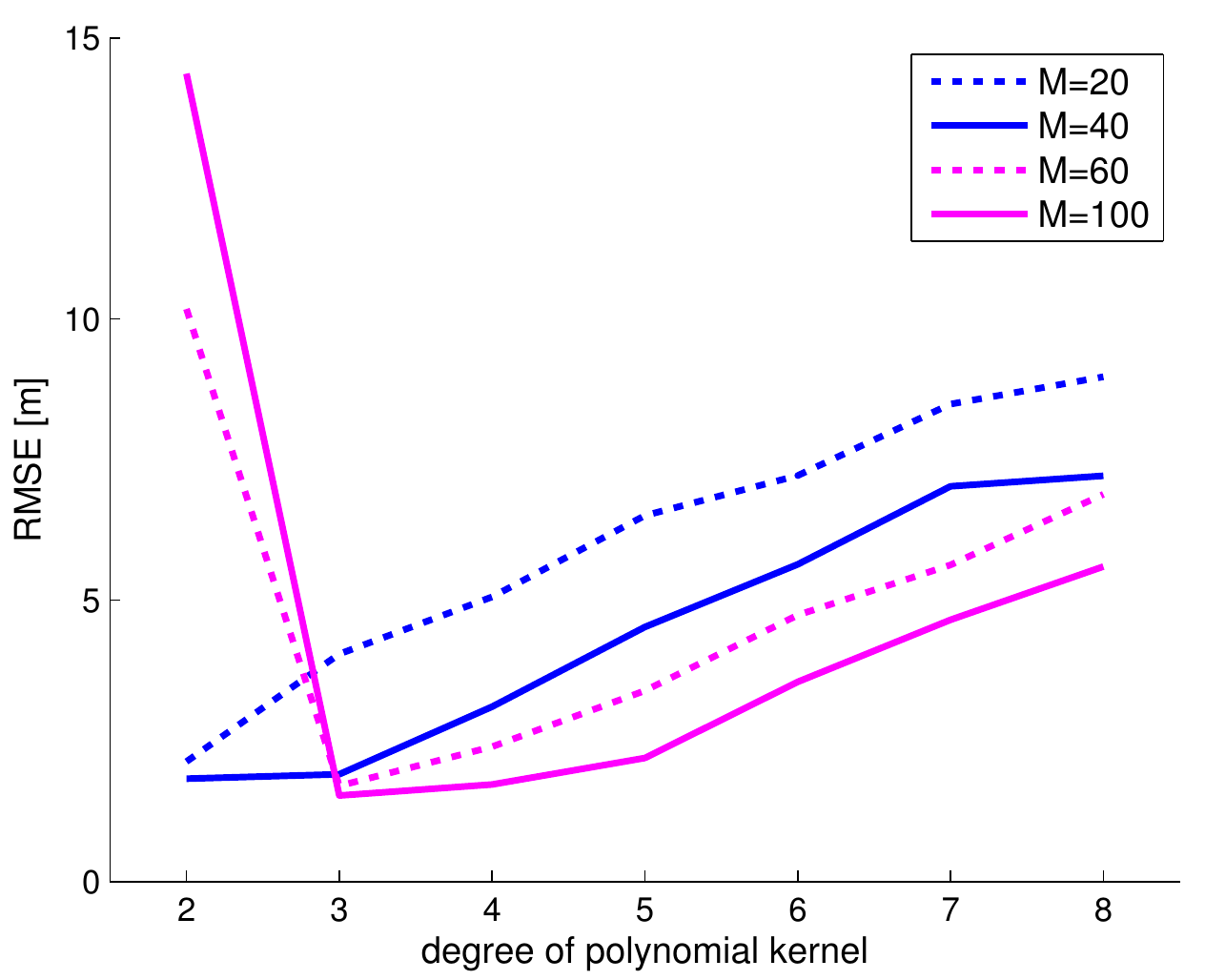}}
\caption{RMSE as a function of degree of polynomial kernel ($c$) for different number of retained principal components ($M$).}
\label{fig:rmseDegreePoly}
\end{figure}

In order to empirically determine appropriate values of $c$ and $M$,
we analyze root-mean-square error (RMSE) of the kPCA range estimates
as a function of degree $c$, for different numbers of retained
principal components $M$ (see Fig. \ref{fig:rmseDegreePoly}). This metric is the most relevant
for many applications, but note that in some cases other metrics (such as median, or 95\%
 percentile) may be also useful.
We see that it is necessary to retain $M \ge 40$ principal
components. We choose $M=60$ principal components, in which case the
best performance is achieved for $c=3$. We also note that a wrong
value of the polynomial degree (e.g., $c=2$) could lead to a
significant loss in performance. In general, too small values of $c$
would lead to under-fitting (i.e., the model is too simple), while too large
values would lead to over-fitting (i.e, the model is too adapted to
training samples).

\begin{figure}[!t]
\centerline{
\subfloat[]{\includegraphics[width=1\columnwidth]{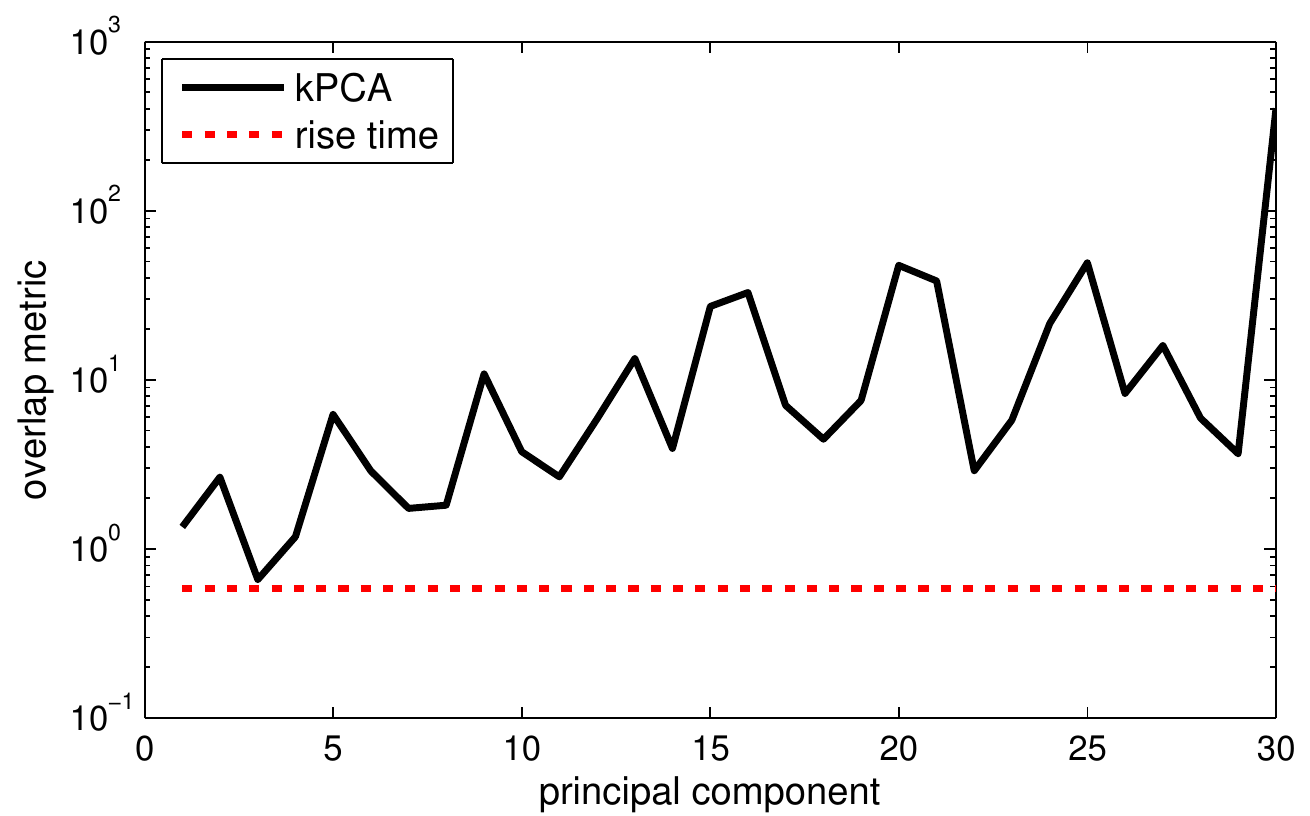}\label{fig:overlapPC}}
}
\centerline{
\subfloat[]{\includegraphics[width=1\columnwidth]{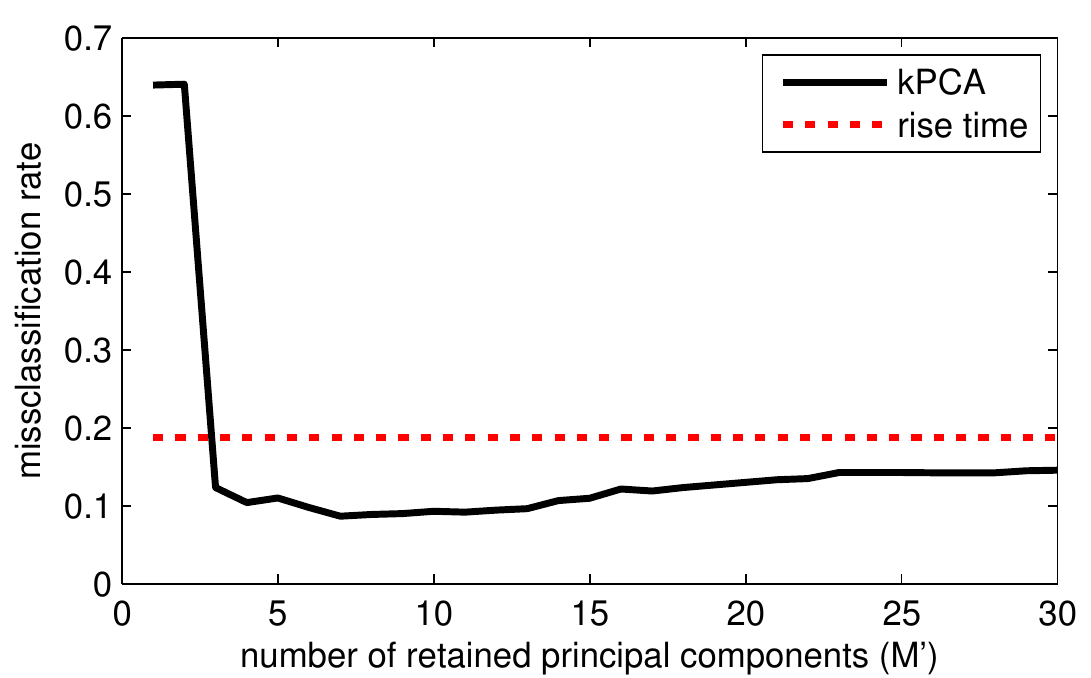}\label{fig:missRateM}}
}
\caption{NLOS identification using kPCA and rise time: (a) overlap metric for different principal components, and (b) misclassification rate as a function of $M'$.}
\label{fig:overlapMissRate}
\end{figure}

Regarding NLOS identification based on kPCA (needed for kPCA+ and
kPCA+GPR), we analyze \textit{overlap metric} \cite{Savic2014tw}, which represents a measure of the overlap of two
distributions considering only their means and variances (defined as
$\sqrt{\sigma _{L,{y_i}}\sigma _{N,{y_i}}}/\left| \mu _{L,{y_i}} - \mu
_{N,{y_i}} \right|$ for $i$-th principal component), and
\textit{misclassification rate} (defined as $N_{miss}/N$, where
$N_{miss}$ is the number of misclassifications). As we can see in
Fig. \ref{fig:overlapPC}, the overlap between LOS and NLOS
distributions have an oscillating behavior (as a function of $i$) and
it is minimal for $i=3$. If we use only this principal component, we
would not be able to perform better than the NLOS identification based
on rise time. We also note an increasing tendency with $i$, since the
higher principal components include little information (i.e., have
small eigenvalues). Therefore, it is reasonable to use the
first $M'$ components as described in Section \ref{sec:kpcagp}.\footnote{Using only
  the local minimums is not recommended since they may change, especially in case of offline training.} In
Fig. \ref{fig:missRateM}, we can see that using kPCA for NLOS
identification, instead of the approach based on rise time, can
significantly reduce the misclassification rate with only $M' \ge 3$
principal components (we choose $M' = 4$ for further analyses). On the
other hand, the misclassification rate never reaches zero, which means
that a soft-decision ranging algorithm is still required. The increasing 
tendency for $M' > 6$ is probably caused by the fact that the uncorrelated principal components 
are not necessarily independent.

\subsection{Performance analysis}\label{sec:perf}

\begin{figure}[!t]
\centerline{
\includegraphics[width=1\columnwidth]{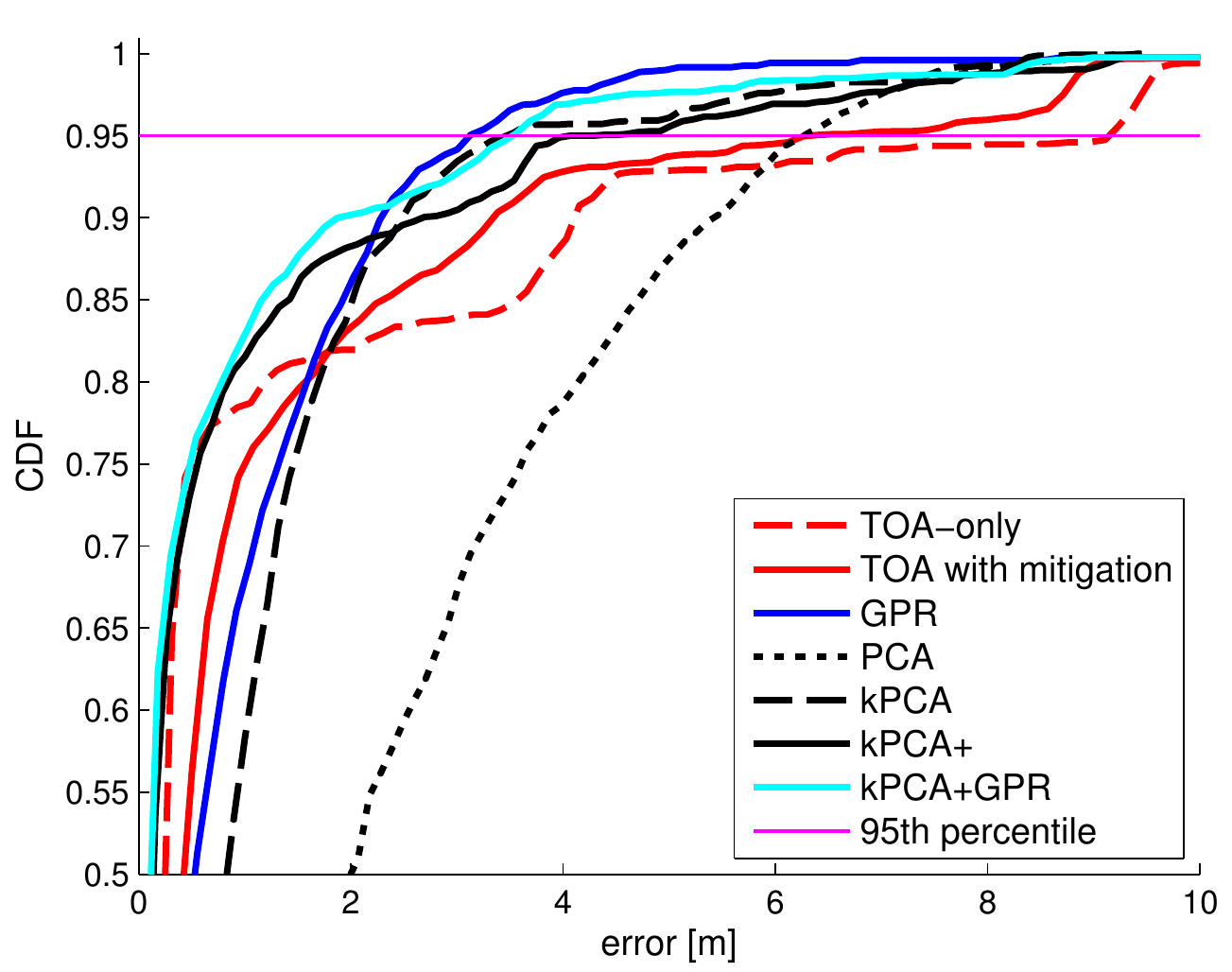}}
\caption{CDF of the ranging error for all considered algorithms. The results are based on 1800 test samples (of which $75\%$ are LOS and 25\% are NLOS).}
\label{fig:cdfErrorCompare}
\end{figure}

Given the previous models, our goal is to analyze the ranging error of
the proposed algorithms and compare it with TOA-based methods and
GPR. Therefore, we analyze the cumulative distribution function (CDF)
of the ranging error, shown in Fig. \ref{fig:cdfErrorCompare}. We
observe the following:
\bi
\item The TOA-only method provides very accurate results for about 77\% of the samples (i.e., for all LOS and few NLOS samples). However, NLOS samples can cause a huge (up to 10 m) error. TOA with mitigation can reduce this error by about 1 m, but it performs slightly worse for LOS samples.

\item The GPR method is the best approach for any percentile above the 90th, but it performs worse than TOA-based methods for the lower percentiles (i.e., mostly LOS samples). That can be explained by the fact that the GPR performs direct ranging using a combination of LOS and NLOS samples (without NLOS identification).

\item The kPCA method performs slightly worse ($\sim 0.4$ m) than GPR for any percentile, due to the approximations made in \eqref{eq:proj-distance}. However, this method is about $N/M=30$ times faster than GPR. For the sake of completeness, we also show the performance of kPCA with the linear kernel (i.e., PCA), which provides a very poor performance, as expected.

\item Both kPCA+ and kPCA+GPR methods, thanks to the NLOS identification,  achieve the performance of the TOA-only method for LOS samples. For NLOS samples, they perform just slightly worse ($\sim$ 0.5-1 m) than GPR. Therefore, these two methods inherit the best characteristics of both state-of-the-art methods.

\ei
In summary, according to our experiments, kPCA+GPR (which is slightly more accurate than kPCA+), is more appropriate method for ranging than other considered state-of-the-art methods.

\begin{figure}[!t]
\centerline{
\subfloat[]{\includegraphics[width=1\columnwidth]{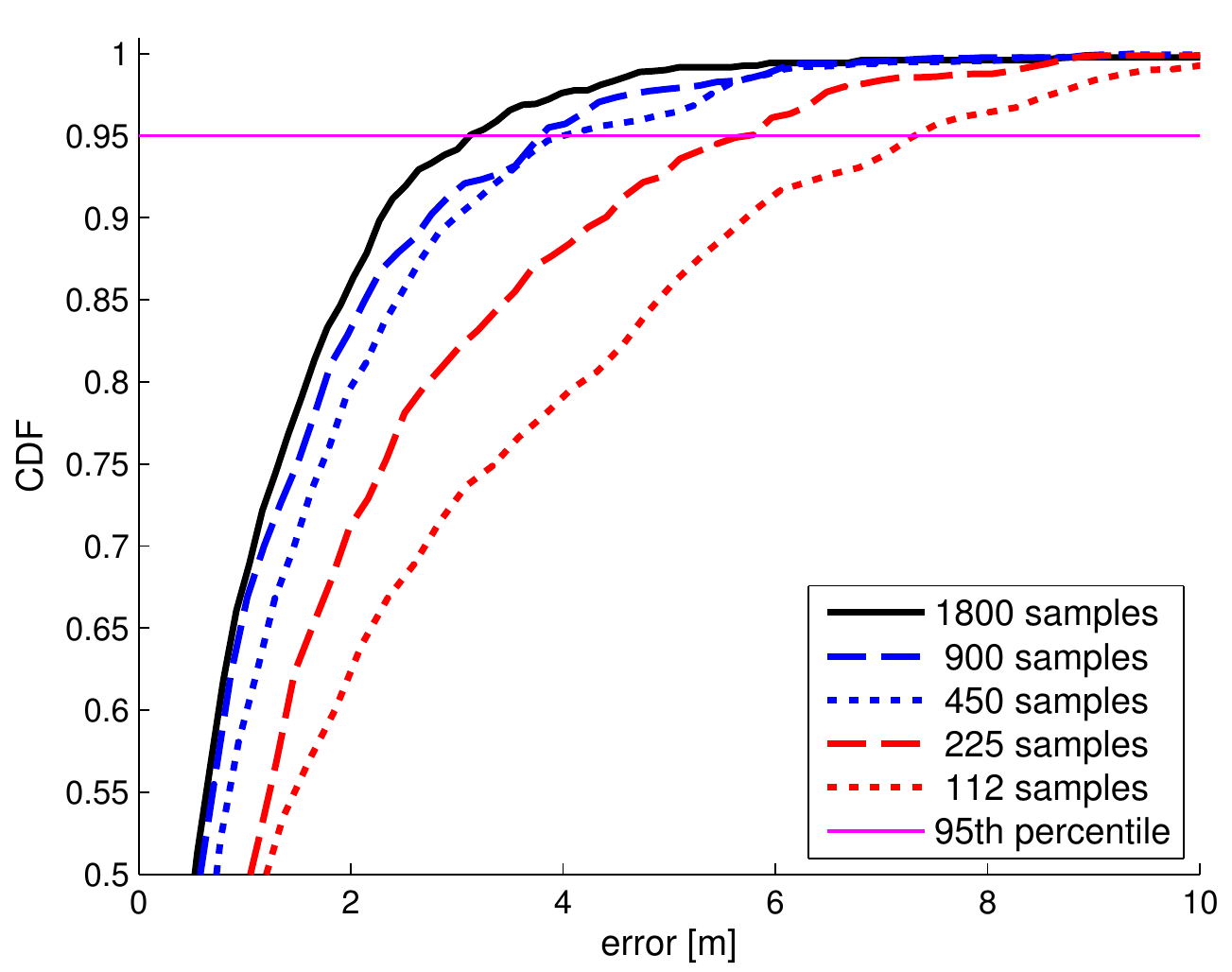}\label{fig:GPRsamples}}
}
\centerline{
\subfloat[]{\includegraphics[width=1\columnwidth]{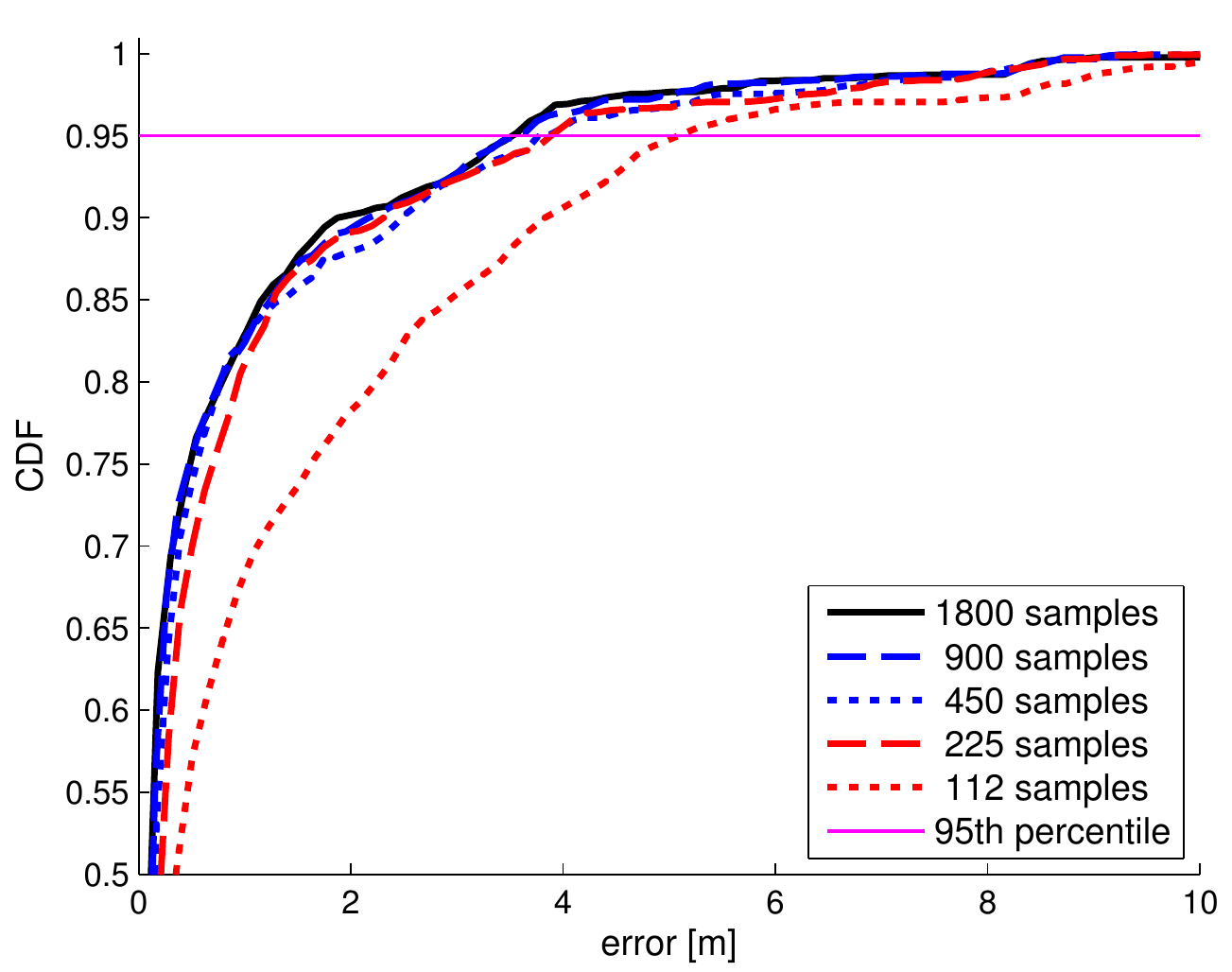}\label{fig:kPCAGPRsamples}}
}
\caption{CDF of the ranging error for different number of the training samples ($N$): (a) GPR, and (b) kPCA+GPR. The ratio between LOS and NLOS samples is preserved in all considered cases.}
\label{fig:trainingSamples}
\end{figure}

The previous analysis assumed that many training samples are
available. We also analyze the effect of the number of training
samples ($N$) on ranging accuracy, for two representative methods: GPR
and kPCA+GPR. The results are shown in
Fig. \ref{fig:trainingSamples}. As we can see, decreasing $N$ leads to
reduced performance of GPR, so we need at least 450 samples to keep
its performance reasonable. On the other hand, the performance of
kPCA+GPR is nearly constant for $225\le N \le 1800$. The reason for
somewhat remarkable behavior is twofold: i) classification is a
discrete problem, in contrast to regression, ii) the kPCA+GPR
estimates partly depend on TOA (see \eqref{eq:dist-est-kpca+}), which
does not require any training samples. Therefore, an additional
advantage of kPCA+GPR is that it requires fewer training
samples, and consequently, it will be much faster.
Moreover, we can observe that, for $100\le N \le 1500$ training samples, kPCA+GPR performs the best, while for $N>1500$, GPR provides the best performance. However, for very few training samples ($N<100$), both algorithms perform worse than TOA with mitigation (comparisons are done at 95th percentile). Consequently, a hybrid algorithm switching from one technique to another, depending on $N$, is also a reasonable option. In that case, the switching values depend on the environment, and can be found empirically.

\begin{figure}[!t]
\centerline{
\includegraphics[width=1\columnwidth]{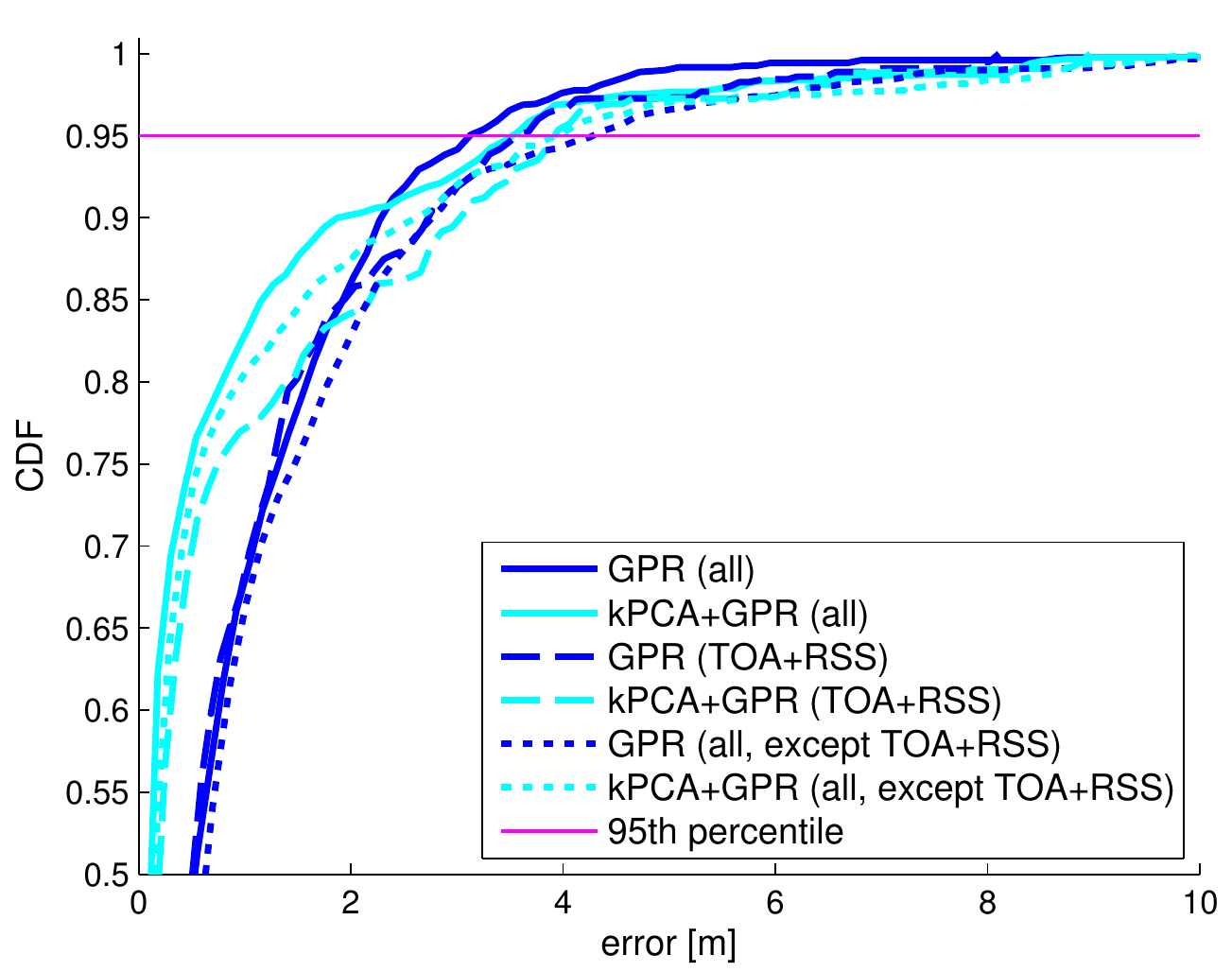}}
\caption{CDF of the ranging error for different subsets of channel parameters.}
\label{fig:cdfSubsetFeat}
\end{figure}

Finally, we analyze how different subsets of channel parameters affect the ranging 
performance. More specifically, we compare GPR and kPCA+GPR ranging methods based on
 i) all 8 channel parameters, ii) only TOA and RSS, and iii) all 
channel parameters except TOA and RSS. As we can observe in Fig. \ref{fig:cdfSubsetFeat}, 
it is beneficial to use more channel parameters for ranging. However, the gain is small (about 0.5 m)
due to the partial correlations between these parameters (the correlation coefficients are available in
\cite[Table III]{Savic2014tw}). Another important observation is that, if ranging is performed using all channel 
parameters except TOA and RSS, we can get a similar performance (for higher percentiles) as with ranging using TOA 
and RSS only. That means that ranging can be performed even \textit{without} traditional metrics, such as 
TOA and RSS, which are used in many ranging and positioning algorithms. Therefore, kernel-based machine learning 
methods represent a very powerful toolbox since the same algorithm is able to \textit{adapt} to different input measurements, even if they are related to the output in a complex nonlinear way.

We finish this section by summarizing the discussed characteristics of
all considered methods in Table \ref{table:ranging-charact}.

\begin{table*}[!t]
\caption{A summary of the main characteristics of the considered ranging methods.}\label{table:ranging-charact}
\centering
{\small \btabu{l || c | c | c | c | c }
Ranging method & \btabu {c} accuracy (50th/95th\\  percentile) \etabu & complexity & \btabu {c} required storage\\ (excluding PDP) \etabu & \btabu{c} NLOS\\ identification \etabu & \btabu {c} number of \\training samples \etabu \\\hline\hline
TOA-only 			& 0.3 m / 9.1 m & $\mathcal{O}(1)$ & $\mathcal{O}(1)$ & no & 0\\
TOA with mitigation & 0.5 m / 6.3 m & $\mathcal{O}(1)$ & $\mathcal{O}(1)$ & yes & 0\\
GPR 				& 0.6 m / 3.1 m & $\mathcal{O}(N^2)$ & $\mathcal{O}(N^2)$ & no & large\\
kPCA 				& 0.9 m / 3.5 m & $\mathcal{O}(MN)$ & $\mathcal{O}(N^2)$ & no & large\\
kPCA+ 				& 0.2 m / 4.1 m & $\mathcal{O}(M'N+MN_N)$ & $\mathcal{O}(N^2+N_N^2)$ & yes & small\\
kPCA+GPR 			& 0.2 m / 3.5 m & $\mathcal{O}(M'N+N_N^2)$ & $\mathcal{O}(N^2+N_N^2)$ & yes & small\\
\hline\hline
\etabu}
\end{table*}

\section{Conclusions}\label{sec:conc}

We proposed novel kernel methods for UWB-based ranging, and
tested them using real data from a tunnel environment. All methods are
much faster than the state-of-the-art kernel method (GPR), since they
use only a subset of orthogonal principal components.  Among the
proposed methods, the kPCA+GPR algorithm performs the best, and it
outperforms both GPR and two TOA-based methods.  In addition, compared
to GPR, it requires fewer training samples. In summary, the moderate
complexity and high ranging performance of kPCA+GPR makes this method
useful for many critical applications, including emergency
situations that require good performance within a short time.

There remain many interesting lines for possible future work. Since
the kernel methods are trained to one particular, fixed environment,
an extension to multiple different environments (e.g., by using online
training) could be of high interest. It would also be useful to study
whether GPR could be improved by replacing the Gaussian process with
some other tractable random process. Finally, finding a tractable way to directly
use channel impulse response measurements (i.e., without selection of features) may provide 
us even better ranging performance.

\section*{Acknowledgments}

The authors would like to thank Per \"{A}ngskog and Jos\'{e} Chilo
(University of G\"{a}vle) for their help during the measurement
campaign.

\footnotesize
\bibliographystyle{ieeetr} 
\bibliography{../../../../Publications/vs-refs,../../../../Publications/others}

\begin{IEEEbiography} [{\includegraphics[width=1in,height=1.25in,clip,keepaspectratio]{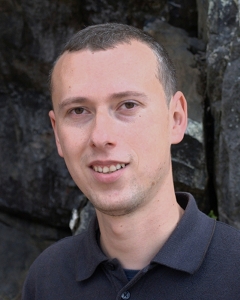}}] {Vladimir Savic}
received the Dipl.Ing. degree in electrical engineering from the University of Belgrade, Belgrade, Serbia, in 2006, and the M.Sc.
and Ph.D. degrees in communications technologies and systems from the Universidad Politecnica de Madrid, Madrid, Spain, in 2009 and 2012, respectively. He was a Digital IC Design Engineer with Elsys Eastern Europe, Belgrade, from 2006 to 2008. From 2008 to 2012, he was a Research Assistant with the Signal Processing Applications Group, Universidad Politecnica de Madrid, Spain. He spent three months as a Visiting Researcher at the Stony Brook University, NY, USA, and four months at the Chalmers University of Technology, Gothenburg, Sweden. In 2012, he joined the Communication Systems (CommSys) Division, Link\"{o}ping University, Link\"oping, Sweden, as a Postdoctoral Researcher. He is co-author of more than 25 research papers in the areas of statistical signal processing and wireless communications. His research interests include localization and tracking, wireless channel modeling, Bayesian inference, machine learning and distributed and cooperative inference over wireless networks.
\end{IEEEbiography}

\begin{IEEEbiography} [{\includegraphics[width=1in,height=1.25in,clip,keepaspectratio]{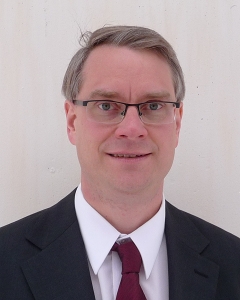}}] {Erik G. Larsson}
received his Ph.D. degree from Uppsala University,
Sweden, in 2002.  Since 2007, he is Professor and Head of the Division
for Communication Systems in the Department of Electrical Engineering
(ISY) at Link\"oping University (LiU) in Link\"oping, Sweden. He has
previously been Associate Professor (Docent) at the Royal Institute of
Technology (KTH) in Stockholm, Sweden, and Assistant Professor at the
University of Florida and the George Washington University, USA.  In the spring of 2015, he was
a Visiting Fellow at Princeton University, USA, for four months.

His main professional interests are within the areas of wireless
communications and signal processing. He has published some 100 journal papers
on these topics, he is co-author of the textbook \emph{Space-Time
Block Coding for Wireless Communications} (Cambridge Univ. Press,
2003) and he holds 15 issued and many pending patents on wireless technology.
He has served as Associate Editor for several major journals, including the \emph{IEEE Transactions on
Communications} (2010-2014) and \emph{IEEE Transactions on Signal Processing} (2006-2010). 
He serves as  chair of the IEEE Signal Processing Society SPCOM technical committee in 2015--2016 and  
as chair of the steering committee for the \emph{IEEE Wireless
Communications Letters} in 2014--2015.  He is the General Chair of the Asilomar Conference
on Signals, Systems and Computers in 2015 (he was Technical Chair in
2012).  He received the \emph{IEEE Signal Processing Magazine} Best Column Award twice, in 2012 and 2014, and he is receiving the IEEE ComSoc Stephen O. Rice Prize in Communications Theory in 2015.
\end{IEEEbiography}

\begin{IEEEbiography} [{\includegraphics[width=1in,height=1.25in,clip,keepaspectratio]{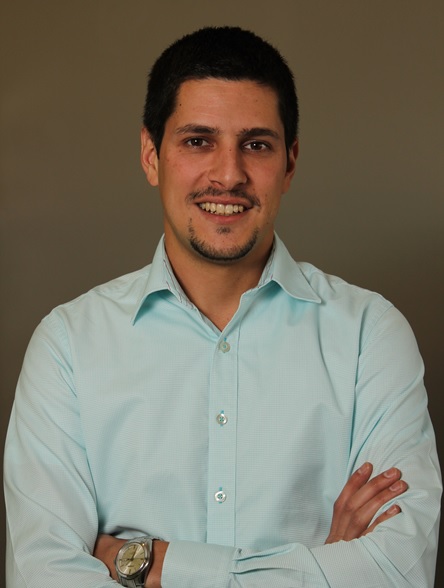}}] {Javier Ferrer-Coll}
received the B.Sc. degree in telecommunication engineering in 2005 and the M.Sc. in 2008 from the Universidad Politecnica de Valencia, Spain. He received his Ph.D in 2014, from the School of Information and Communication, Royal Institute of Technology (KTH), Stockholm, Sweden. From 2009 to 2014, he was employed at the university of G\"avle, Sweden. He is currently employed as system engineer in the communication department at Combitech AB. His research interest is channel characterization in industrial environments, particularly measurement system design to extract the channel characteristics. Moreover, he is also interested in techniques to detect and suppress electromagnetic interferences.
\end{IEEEbiography}

\begin{IEEEbiography} [{\includegraphics[width=1in,height=1.25in,clip,keepaspectratio]{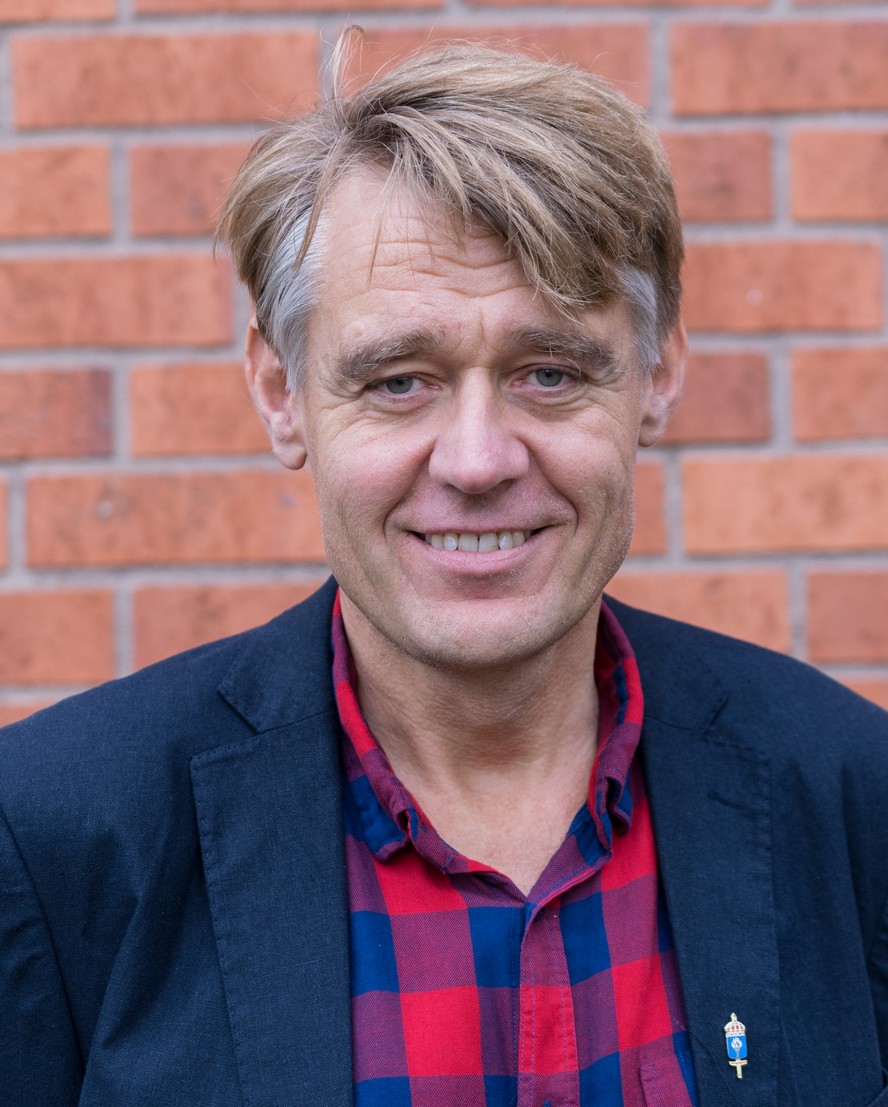}}] {Peter Stenumgaard}
has a Ph.D in radio communications from the Royal Institute of Technology (KTH), Stockholm. He is currently a Research Director and works as Head of the Department of Information Security \& IT architecture at the Swedish Defence Research Agency (FOI) in Link\"oping, Sweden. He has worked as adjunct professor, both at Link\"oping University, Sweden, and the University of G\"avle, Sweden. He has long experience of research for both military and civilian wireless applications and has also been the Director of the graduate school Forum Securitatis (funded by Vinnova) within Security and Crisis Management at Link\"oping University. He worked for several years on the JAS fighter aircraft project with the protection of aircraft systems against electromagnetic interference, lightning, nuclear weapon-generated electromagnetic pulse (EMP) and high-power microwaves (HPM). His research interests are robust telecommunications for military-, security-, safety- and industrial use.
\end{IEEEbiography}

\end{document}